\documentclass[a4paper,conference]{IEEEtran}
\IEEEoverridecommandlockouts
\usepackage{cite}
\usepackage{amsmath,amssymb,amsfonts}
\usepackage{graphicx}
\usepackage{textcomp}
\usepackage{xcolor}
\usepackage{multirow} 
\usepackage{float}  % Add this in the preamble
\usepackage{algorithm}
\usepackage{algpseudocode} 
\usepackage{subcaption}

\def\BibTeX{{\rm B\kern-.05em{\sc i\kern-.025em b}\kern-.08em
    T\kern-.1667em\lower.7ex\hbox{E}\kern-.125emX}}

\begin{document}

\title{Real-Time Obstacle Avoidance for a Mobile Robot Using CNN-Based Sensor Fusion}

\author{\IEEEauthorblockN{1\textsuperscript{st} Lamiaa H. Zain}
\IEEEauthorblockA{
School of Engineering and 
Applied Science  \\
SESC Research Center, MECT Program\\
Nile University\\
Giza, Egypt \\
lzain@nu.edu.eg}

}

\maketitle

% ----------------- Abstract --------------------

\noindent\textbf{Accepted Paper Notice:}  
This work has been accepted for presentation at the 2025 Novel Intelligent and Leading Emerging Sciences  (NILES 2025) Conference. © 2025 IEEE. Personal use of this material is permitted. Permission from IEEE must be obtained for all other uses, including reprinting, republishing, or for creating new works.

\begin{abstract}
Obstacle avoidance is a critical component of the navigation stack required for mobile robots to operate effectively in complex and unknown environments. In this research, three end-to-end Convolutional Neural Networks (CNNs) were trained and evaluated offline and deployed on a differential-drive mobile robot for real-time obstacle avoidance to generate low-level steering commands from synchronized color and depth images acquired by an Intel RealSense D415 RGB-D camera in diverse environments. Offline evaluation showed that the NetConEmb model achieved the best performance with a notably low MedAE of $0.58 \times 10^{-3}$ rad/s. In comparison, the lighter NetEmb architecture, which reduces the number of trainable parameters by approximately 25\% and converges faster, produced comparable results with an RMSE of $21.68 \times 10^{-3}$ rad/s, close to the $21.42 \times 10^{-3}$ rad/s obtained by NetConEmb. Real-time navigation further confirmed NetConEmb's robustness, achieving a 100\% success rate in both known and unknown environments, while NetEmb and NetGated succeeded only in navigating the known environment. 
\end{abstract}


% ------------------ Introduction ------------------

\section{Introduction}

\begin{figure}[!t]
\centering
\includegraphics[width=\columnwidth]{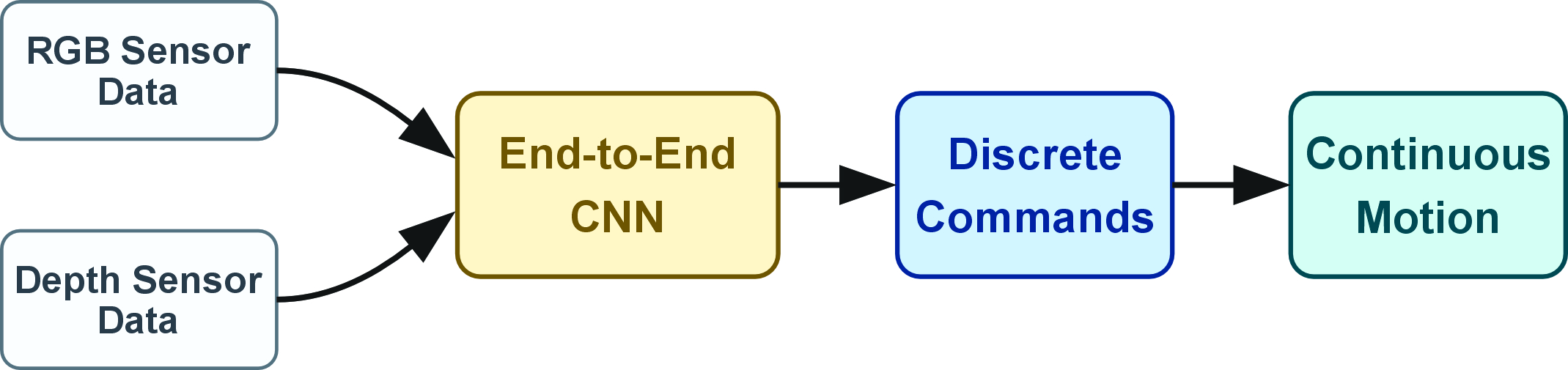}
\caption{End-to-end navigation approach.}
\label{fig:02}
\end{figure}

Mobile robot navigation is essential for tasks such as reaching specific destinations and avoiding obstacles. Navigation strategies can be broadly classified into two categories: classical (conventional) and end-to-end approaches~\cite{xiao2022motion}. Classical navigation relies on sensory inputs such as cameras, Light Detection and Ranging (LiDAR) sensors, and sonar to build an environmental representation (mapping). This representation is processed by a global planner, which determines a global or intermediate goal, and a local planner, which uses this goal along with the environmental map to generate the mobile robot’s steering commands. In contrast, end-to-end navigation directly maps sensory data to low- or high-level steering commands for navigation objectives as shown in Fig.~\ref{fig:02}.

The classic approach, as in~\cite{li2021design}, necessitates numerous phases and a significant amount of engineering effort. As a result, the importance of an end-to-end method emerges, as it allows navigation at a higher degree of semantic abstraction~\cite{sepulveda2018deep} without requiring computation of the robot's location or map generation. Unlike the classic approach, which needs to have a pre-specified goal to reach, the end-to-end approach interacts with the surrounding environment instantly~\cite{lu2022online}.
%%%%%%%%%%%%%%%%%%%%%%%%%%%%%%%%%%%%%%%%%%%%
%\begin{figure}[!t]
%\centering
%\includegraphics[width=\columnwidth]{images/System architecture of the TurtleBot2 with onboard vision, motor control, and remote server communication..PNG}
%\caption{System architecture of the mobile robot with onboard vision, motor control, and remote server communication.}
%\label{fig:01}
%\vspace{1em} % optional small gap
%%%%%%%%%%%%%%%%%%%%%%%%%%%%%%%%%
\vspace{0.1em} % optional small gap
%%%%%%%%%%%%%%%%%%%%%%%%%%%%
Categorized by the input sensory data, end-to-end approaches can be classified into three categories: geometric, which has spatial information about the position and location of surrounding objects, such as data obtained from LiDAR sensors; non-geometric, which pertains to data expressing environmental features like color and texture, such as images from RGB cameras; and hybrid, which is a fusion of both types. In this study, a hybrid approach was utilized, whereby both depth images and color images were considered vital.  

The main goal of this study, based on previous motivation, was to achieve obstacle avoidance in both known and unknown environments using end-to-end imitation learning with sensor fusion of color and depth data, where the system learns from human demonstrations. 
%Herein, ”known environments” refer to the real environments in which the robot was exposed during data collection, while "unknown environments" denote previously unseen locations with different spatial configurations and object arrangements.%
Three custom CNNs were investigated and compared to find the most acceptable one with the best evaluation metrics. In addition, all of them were applied to a mobile robot to test online real-time navigation. 

Several challenges were encountered during this work. One was the use of the Intel RealSense D415 sensor, which integrates both RGB and depth sensing in a single compact unit, unlike previous studies that relied on separate LiDARs and RGB cameras. Another challenge was managing the relatively large dataset (1.745 GB), which was collected across diverse environments with varying object types to ensure robust training and testing of all CNN models. The last challenge was the real-time implementation of the three networks for mobile robot navigation in both known and unknown environments.

% ----------------- Related Work --------------------

\section{Related Work}

This section presents a concise overview of relevant research for the current study.

Recurrent Neural Networks (RNNs) and 3D-CNNs were used by~\cite{park2020vision} for obstacle avoidance in simulated Unmanned Aerial Vehicles (UAVs), using visual images and flight data as inputs. Similarly,~\cite{lee2021deep} employed a Faster Region-based CNN to detect and avoid tree trunks in low-altitude UAVs using a single camera.

In another imitation learning setup,~\cite{codevilla2018end} applied conditional imitation learning for road navigation using images from three cameras, along with robot speed, position, and goal information to generate high-level steering commands. Similarly,~\cite{sepulveda2018deep} used CNN-based supervised imitation learning for navigation across diverse environments, using visual observations with subgoals as inputs and vehicle velocities as outputs.

 In~\cite{9521224}, a CNN architecture was proposed to predict motion control commands for mobile robot navigation using information from both LiDAR and relative target position, while~\cite{yan2022mapless} used imitation learning for obstacle avoidance relying on LiDAR data through their contributions to the loss function and data augmentation.

Some studies have focused on scenarios involving a moving goal, such as lane tracking and dynamic obstacle avoidance. For example,~\cite{kim2022end} used Long-term Recurrent Convolutional Networks (LRCNs) to infer the steering angle and speed by providing the vehicle’s speed and color images as input to the network. Similarly,~\cite{nair2020robotic} used Long Short-Term Memory (LSTM) networks for optimal path generation with raw images and Dijkstra-generated paths as inputs, while~\cite{bai2018deep} employed spatiotemporal LSTM for autonomous vehicle motion control. In another study,~\cite{9144508} also used LSTM to steer a wheeled mobile robot. Finally,~\cite{siva2019robot} introduced a novel approach that combines inverse reinforcement learning (IRL) with feature representation to adapt the robot’s motion to various types of unstructured terrain.

This study builds upon and extends our previous work~\cite{zain2025arxiv}. It introduces the NetEmb architecture, a lightweight CNN with approximately one-quarter the number of trainable parameters compared to the previously best-performing network, NetConEmb, while employing the same data fusion mechanism (concatenation). Alternative fusion strategies, such as weighted averaging and attention mechanisms, were considered, but
concatenation was chosen as an intermediate fusion technique for its simplicity, computational
efficiency, and proven effectiveness in preserving spatial information from both modalities. A key distinction of this work is the real-time deployment of all three networks on a mobile robot to evaluate their navigation performance in both known and unknown environments.

% ----------------- Experimental Setup --------------------

\section{Experimental Setup}

The hardware used in this research was illustrated in our previous work~\cite{zain2025arxiv}. It has its design inspired by TurtleBot2, a mobile robot by Yujin Robotics in Korea.

Ubuntu 18.04.6 was used as the operating system for both the Remote Server and the on-robot computer (Nvidia Jetson Nano). PyTorch was used as a Python framework to train and test the designed architectures of all CNNs. Robots Operating System (ROS) was used in both phases of data collection and online navigation to facilitate communication between different nodes like the Camera, the Arduino, and the Remote Server.

\begin{figure}[!b]
\centering
\includegraphics[width=\columnwidth]{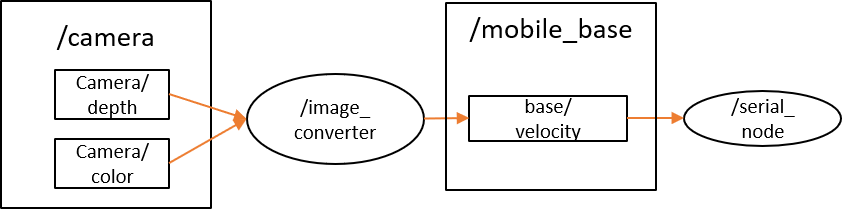}
\caption{RQT graph generated during online navigation.}
\label{fig:04}
\end{figure}

\begin{figure}[!t]
\centering
\includegraphics[width=0.8\columnwidth]{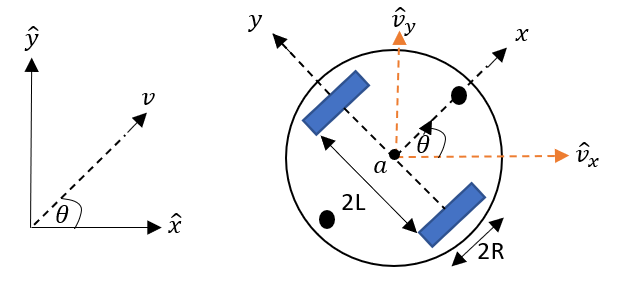}
\caption{Geometrical representation of the differential drive mobile robot showing coordinate frame definitions.}
\label{fig:05}
\end{figure}

The ROS Qt Tool (RQT) graph generated by ROS during online navigation is shown in Fig.~\ref{fig:04}, where the names in ellipses represent nodes, and those in rectangles represent topics. During online navigation, once the mobile robot is started, the camera captures color and depth images frame by frame (as ROS topics). These images are sent to the image converter node to generate color and depth data, which serve as raw inputs to the model. The model then outputs low-level steering commands (angular velocity), which are published to the Arduino serial node. At this stage, the kinematic equations are applied to convert the robot's linear velocity (fixed at 0.1 m/s) and angular velocity, $\dot{x}$ and $\dot{\theta}$, into the left and right motor velocities, $\dot{\varphi}_{L}$ and $\dot{\varphi}_{R}$. Equation~\eqref{Eq:7} represents the kinematic model relative to the robot frame \{${x},${y}\}, while Equation~\eqref{Eq:8} corresponds to the global frame \{$\hat{x},\hat{y}$\}, as illustrated in Fig.~\ref{fig:05}.

%%%%%%%%%%%%%%%%%%%%%%%
%%%%%%%%%%%%%%%%%%%%%%%

%%%%%%%%%%%%%%%%%%%%%%%
\begin{figure}[!b]
\centering
\includegraphics[width=\columnwidth]{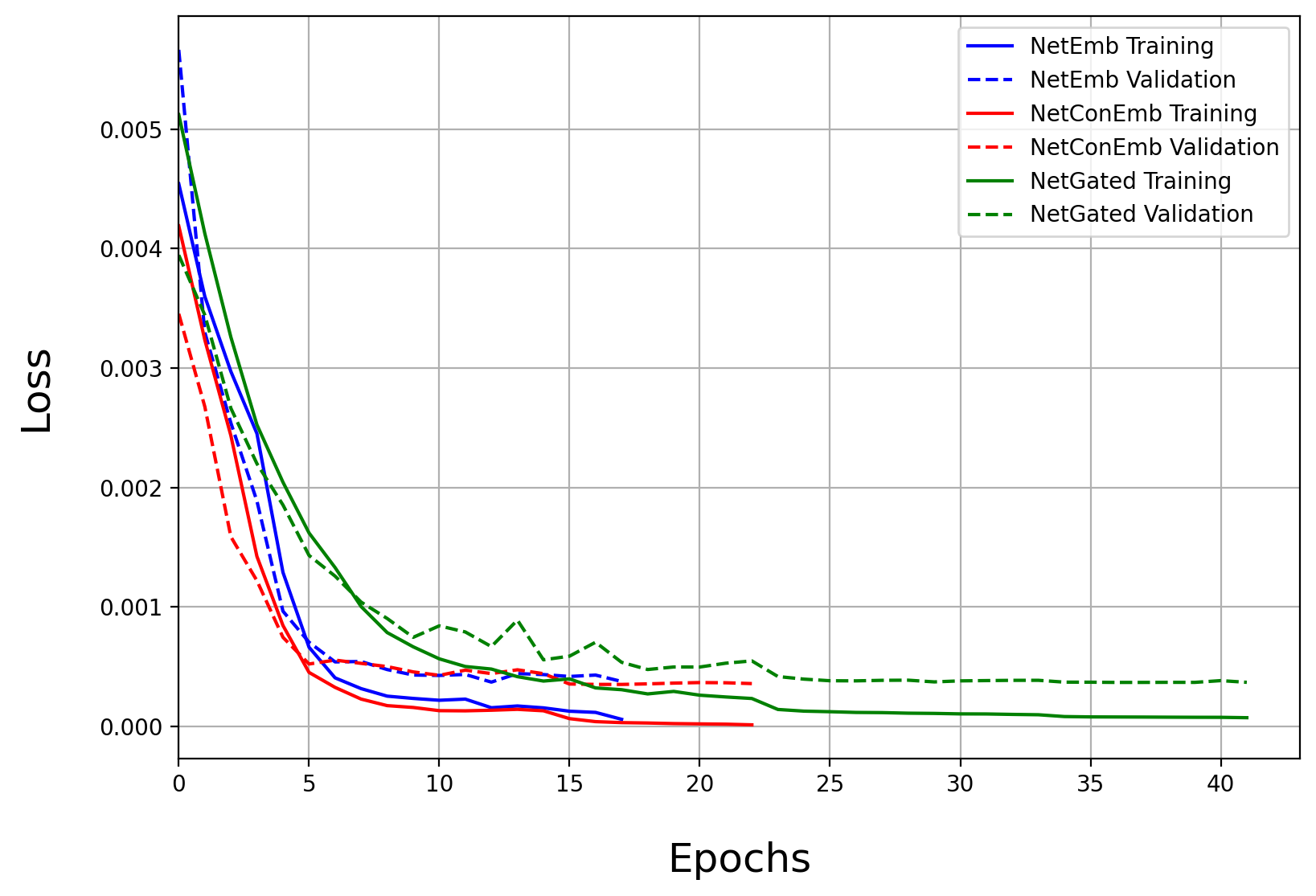}
\caption{Training and validation loss comparison of the three CNN architectures with early stopping applied, demonstrating convergence characteristics and relative performance.}
\label{fig:07}
\end{figure}
%%%%%%%%%%%%%%%%%%%%%%%

%%%%%%%%%%%%%%%%%%%%%%%

{\small
\begin{equation}
\begin{aligned} 
\left[\begin{array}{c}
\dot{x} \\
\dot{y} \\
\dot{\theta}
\end{array}\right]=\left[\begin{array}{cc}
\frac{R}{2} & \frac{R}{2} \\
0 & 0 \\
\frac{R}{2 L} & -\frac{R}{2 L}
\end{array}\right]\left[\begin{array}{l}
\dot{\varphi}_{R} \\
\dot{\varphi}_{L}
\end{array}\right]
\end{aligned}
\label{Eq:7}
\end{equation}
}

{\small
\begin{equation}
\begin{aligned}
\left[\begin{array}{c}
\dot{\hat{x}} \\
\dot{\hat{y}} \\
\dot{\hat{\theta}}
\end{array}\right]=\left[\begin{array}{cc}
\frac{R}{2} \cos \theta & \frac{R}{2} \cos \theta \\
\frac{R}{2} \sin \theta & \frac{R}{2} \sin \theta \\
\frac{R}{2 L} & -\frac{R}{2 L}
\end{array}\right]\left[\begin{array}{c}
\dot{\varphi}_{R} \\
\dot{\varphi}_{L}
\end{array}\right]
\end{aligned}
\label{Eq:8}
\end{equation}
}

{\small
\begin{equation}
\dot{\varphi}_{R} = \frac{\dot{x} + \dot{\theta} L}{R}, \quad
\dot{\varphi}_{L} = \frac{\dot{x} - \dot{\theta} L}{R}
\label{Eq:9}
\end{equation}
}

%%%%%%%%%%%%%%

\begin{figure*}[!t]
\centering
\includegraphics[width=\textwidth]{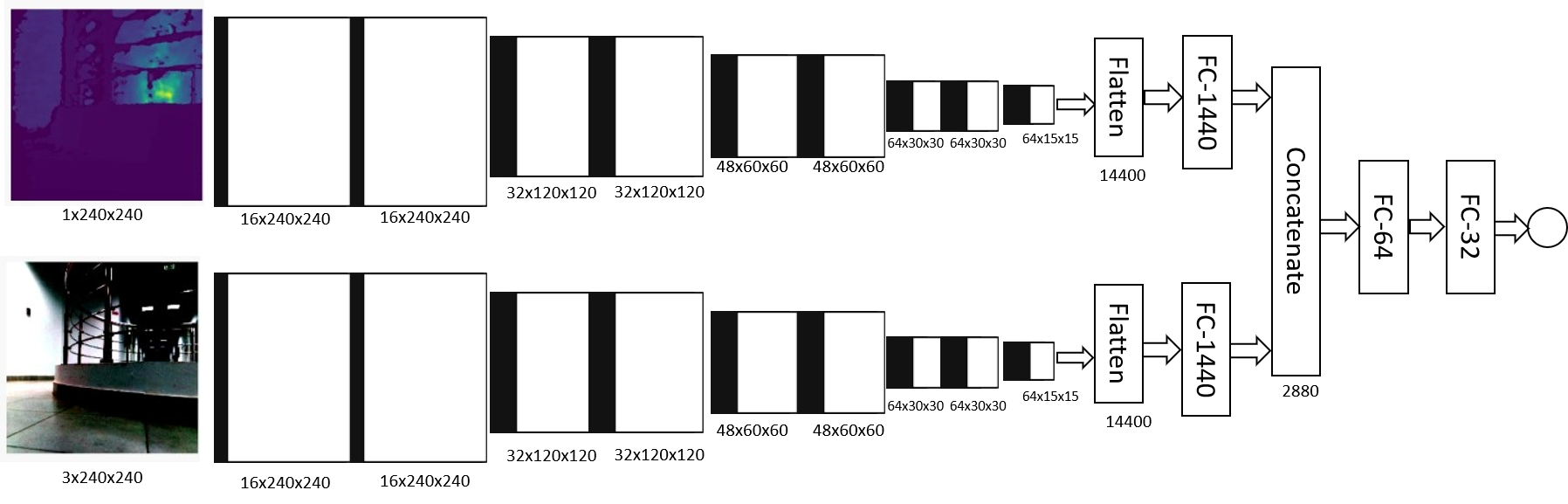}
\caption{NetEmb architecture showing the late-fusion approach with embedded feature concatenation, resulting in reduced parameter count while maintaining performance characteristics.}
\label{fig:06}
\end{figure*}

%----------------- Methodology -------------

\section{Methodology}

\subsection{Proposed End-To-End Architecture}

In our previous study,~\cite{zain2025arxiv}, the architectures of two networks, NetConEmb and NetGated, were investigated, with NetConEmb demonstrating better performance. In the current study, a third CNN architecture, NetEmb (Fig.~\ref{fig:06}), is used and compared against the previous two. NetEmb architecture was adopted from earlier work ~\cite{patel2019deep}, with adaptations made for the depth sensor and changes to the dimensions of the input images. NetEmb closely resembles NetConEmb in architecture, sharing identical feature extraction layers and employing the same data fusion strategy via concatenation. However, NetEmb reduces the number of trainable parameters by performing concatenation after generating an embedded feature vector at the fully connected layer FC-1440. In contrast, NetConEmb performs concatenation immediately after feature extraction. As a result, NetEmb has 21,040,049 trainable parameters, approximately 25\% fewer than those of NetConEmb. This reduction improves memory efficiency and contributes to faster convergence. As shown in Fig.~\ref{fig:07}, NetEmb completes training at epoch 18, while NetConEmb converges at epoch 23, with both achieving similar final training and validation losses.

\subsection{Models' Training and Optimization Stage}

The dataset, loss function, and optimizer used in this study are the same as those described in our previous work~\cite{zain2025arxiv}. However, the learning rate (LR) was selected using a grid search algorithm. Several candidate LRs were evaluated by training the model for only five epochs each, and the LR yielding the lowest training loss was chosen as the initial LR, as shown in Table~\ref{table1}. To avoid stagnation, a scheduler reduced the LR by a factor of 0.2 if the validation loss plateaued for three consecutive epochs. Additionally, early stopping was applied to terminate training if no improvement in validation loss was observed over five consecutive epochs. All training experiments were conducted using an NVIDIA RTX 4080S GPU. As shown in Fig.~\ref{fig:07}, NetEmb and NetConEmb have similar learning curves, while NetGated, trained for 42 epochs, shows higher training and validation losses.
%%%%%%%%%%%%%%%%%%%%%%%

\begin{table}[!b]
    \renewcommand{\arraystretch}{1.3}
    \caption{Grid Search results for initial learning rate selection.}
    \label{table1}
    \centering
    \resizebox{\columnwidth}{!}{
        \begin{tabular}{l c c c c c}
            \hline\hline \\[-3mm]
            \textbf{\shortstack{Network /\\ LR}} & 
            \textbf{1e-06} & 
            \textbf{3e-06} & 
            \textbf{3e-05} & 
            \textbf{1e-04} & 
            \textbf{3e-04} \\[1.6ex] \hline
            NetEmb (current) & 0.004212 & 0.003807 & 0.002483 & \textbf{0.002042} & 0.002245 \\
            NetConEmb~\cite{zain2025arxiv} & 0.004170 & 0.004128 & 0.001919 & 0.001897 & \textbf{0.001543} \\
            NetGated~\cite{zain2025arxiv} & 0.004328 & 0.004327 & \textbf{0.002787} & 0.004444 & 0.004257 \\[1.4ex]
            \hline\hline
        \end{tabular}
    }
\end{table}
%%%%%%%%%%%

%----------------- Results -------------

\section{Results}

\subsection{Network Performance Comparison}

To evaluate the performance of the three networks prior to online deployment, four metrics were computed on the test dataset: Mean Absolute Error (MAE), Root Mean Squared Error (RMSE), Median Absolute Error (MedAE), and Variance Score (VS). The first three metrics are in rad/sec and the final one is unitless. The corresponding results are presented in Table~\ref{table2}. As observed, NetEmb and NetConEmb yield comparable RMSE and MAE values. However, NetConEmb demonstrates a slight advantage, particularly with a substantially lower MedAE, approximately half that of NetEmb, suggesting that its predictions are more consistently close to the ground truth, with smaller errors in the majority of test cases.
%%%%%%%%%%%%%%%%%%%%%%%%%%
\begin{table}[!b]
    \renewcommand{\arraystretch}{1.3}
    \caption{Performance metrics of the trained networks on the testing dataset.}
    \label{table2}
    \centering
    \resizebox{\columnwidth}{!}{
        \begin{tabular}{l c c c c}
            \hline\hline \\[-3mm]
            \textbf{Network} & 
            \textbf{\shortstack{MAE \\ $\times 10^3$}} & 
            \textbf{\shortstack{RMSE \\ $\times 10^3$}} & 
            \textbf{\shortstack{MedAE \\ $\times 10^3$}} & 
            \textbf{VS} \\[1.6ex] \hline
            NetEmb (current) & 6.58 & 21.68 & 1.19 & 0.94 \\
            NetConEmb~\cite{zain2025arxiv} & 6.04 & 21.42 & 0.58 & 0.94 \\
            NetGated~\cite{zain2025arxiv} & 9.51 & 22.88 & 3.49 & 0.93 \\[1.4ex]
            \hline\hline
        \end{tabular}
    }
\end{table}

%%%%%%%%%%%%%%%%%%%%%%%%%%%%%%%%%%%%%

\subsection{Ablation Study Results}

An ablation study was performed to evaluate the impact of each individual modality on each network's performance. For each of the three architectures, separate training runs were conducted using only one modality, either color or depth, while the other modality was set to zero to simulate sensor failure. This approach isolates the contribution of each sensor input and enables analysis of the network's robustness under partial input conditions. The results of this study are summarized in Table~\ref{table3}.

\begin{table}[!t]
    \renewcommand{\arraystretch}{1.3}
    \caption{Ablation study results for single-modality inputs across all network architectures.}
    \label{table3}
    \centering
    \resizebox{\columnwidth}{!}{
        \begin{tabular}{l l c c c c}
            \hline\hline \\[-3mm]
            \textbf{Network} & 
            \textbf{Input Type} & 
            \textbf{{\shortstack{MAE \\ $\times 10^3$}}} & 
            \textbf{{\shortstack{RMSE \\ $\times 10^3$}}} & 
            \textbf{{\shortstack{MedAE \\ $\times 10^3$}}} & 
            \textbf{VS} \\[1.6ex] \hline
            \multirow{2}{*}{NetEmb (current)} & Color Images & 6.05 & 18.76 & 1.60 & 0.96 \\
                                    & Depth Images & 9.38 & 28.94 & 1.89 & 0.90 \\ \hline
            \multirow{2}{*}{NetConEmb~\cite{zain2025arxiv}} & Color Images & 4.81 & 19.90 & 0.32 & 0.95 \\
                                       & Depth Images & 9.71 & 32.18 & 0.98 & 0.87 \\ \hline
            \multirow{2}{*}{NetGated~\cite{zain2025arxiv}} & Color Images & 6.47 & 20.90 & 1.36 & 0.95 \\
                                      & Depth Images & 10.38 & 30.61 & 2.20 & 0.89 \\[1.4ex]
            \hline\hline
        \end{tabular}
    }
\end{table}

%%%%%%%%%%%%%%%%%%%%%%%%%%%%%%%%%
The results in Table~\ref{table3} indicate that, across all architectures, training with color images alone yielded lower error metrics compared to training with depth images alone. In particular, NetConEmb achieved the best performance with color input, exhibiting the lowest MAE and a notably small MedAE, suggesting highly consistent predictions on the test dataset. When relying solely on depth data, NetEmb and NetConEmb produced comparable results, with a slight advantage for NetEmb, whereas NetGated exhibited the highest error. However, it is important to emphasize that these results are limited to offline evaluation.

\subsection{Real-Time Navigation Performance}

\subsubsection{Multi-Environment Testing}

\begin{figure}
  \centering
  \begin{subfigure}[b]{0.48\linewidth}
    \includegraphics[width=\linewidth]{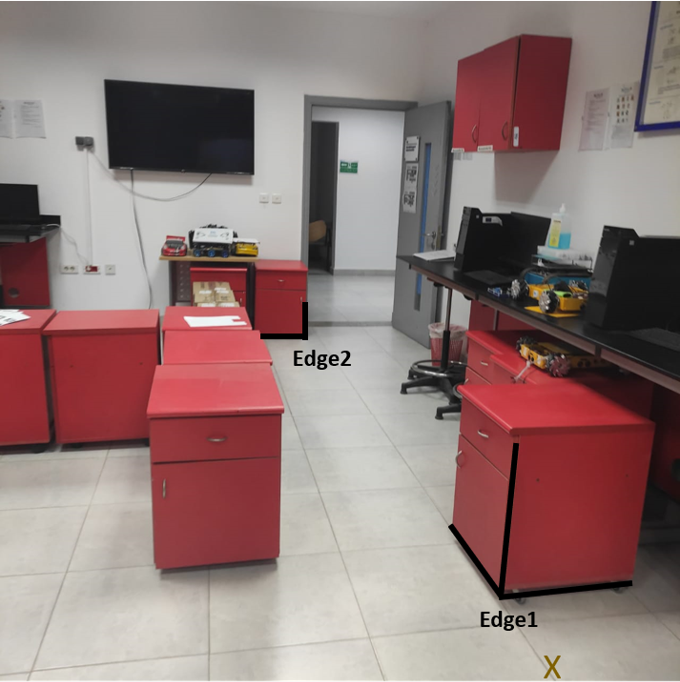}
    \caption{}
  \end{subfigure}
  \hfill
  \begin{subfigure}[b]{0.48\linewidth}
    \includegraphics[width=\linewidth]{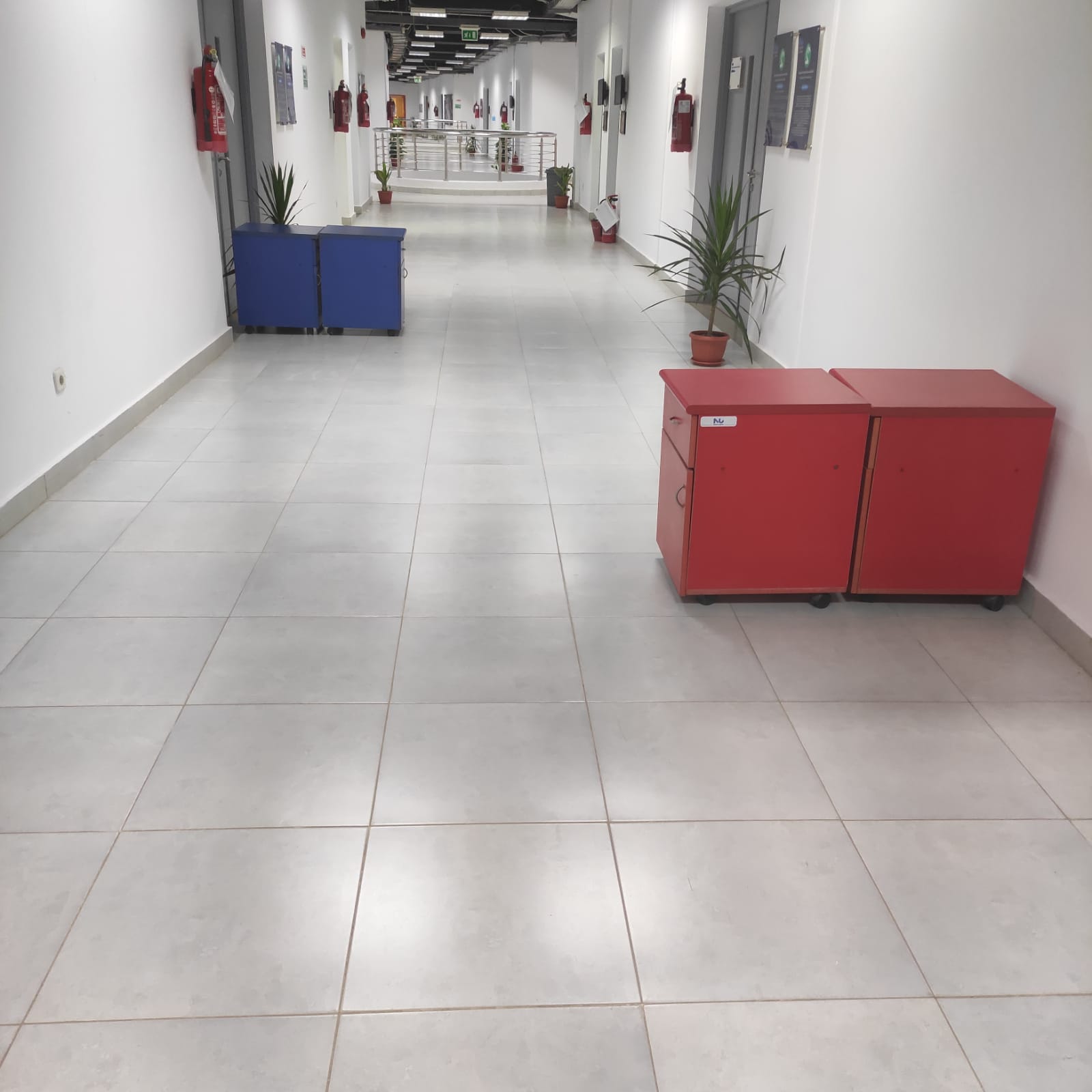}
    \caption{}
  \end{subfigure}
  \caption{Experimental environments used for testing real-time navigation: (a) unknown lab environment, (b) known corridor environment.}
  \label{fig:obstacle_envs}
\end{figure}

\begin{figure}
  \centering
  \begin{subfigure}[b]{0.48\linewidth}
    \includegraphics[keepaspectratio,height=6cm]{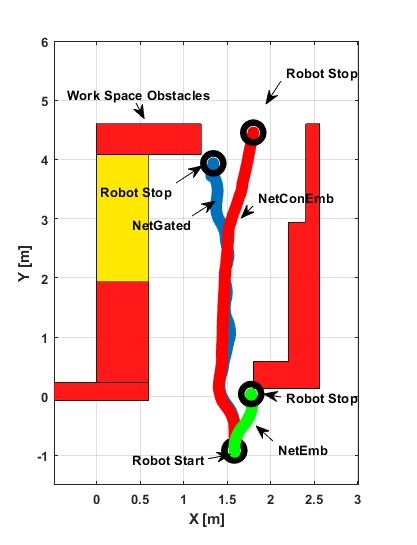}
    \caption{}
  \end{subfigure}
  \hfill
  \begin{subfigure}[b]{0.48\linewidth}
    \includegraphics[keepaspectratio,height=6cm]{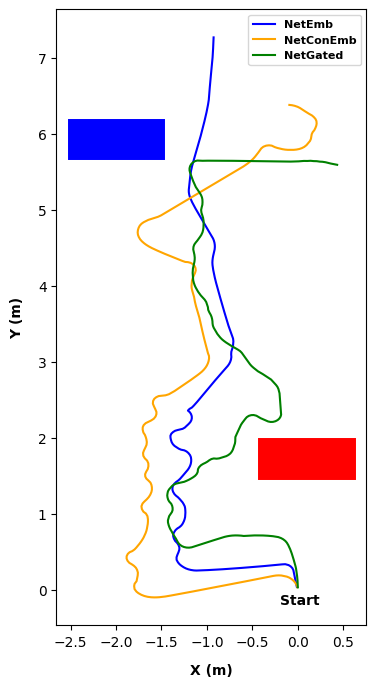}
    \caption{}
  \end{subfigure}
  \caption{Navigation paths in unknown environments (a) and known environments (b) using information from both modalities.}
  \label{fig:paths}
\end{figure}

To evaluate real-time navigation performance, two experimental setups were used. The first involved an unseen environment, a narrow laboratory lane that the robot had not previously encountered, Fig.~\ref{fig:obstacle_envs}a, while the second was a known corridor environment in which the dataset was collected, Fig.~\ref{fig:obstacle_envs}b. Each network was tested in both environments using fused color and depth data, with three trials conducted to support consistent conclusions. The resulting navigation paths are plotted as shown in Fig.~\ref{fig:paths}a and Fig.~\ref{fig:paths}b. During navigation, the on-robot computer (NVIDIA Jetson Nano) published the linear and angular velocities via ROS topics approximately every 0.2 seconds. NetConEmb successfully achieved obstacle avoidance in both environments with a successful navigation rate of 100\%, whereas NetGated and NetEmb encountered difficulties in the unfamiliar setting but succeeded in navigating the known environment, where each model achieved successful navigation in 3 out of 6 cases (50\%) in both environments.
%%%%%%%%%%%%%%%%%%%%%%%%%%%
%%%%%%%%%%%%%%%%%%%%%%%%%%%%%%%%%%

\subsubsection{Sensor Failure Analysis}

\begin{figure}
  \centering
  \begin{subfigure}[b]{0.48\linewidth}
    \includegraphics[keepaspectratio,height=6cm]{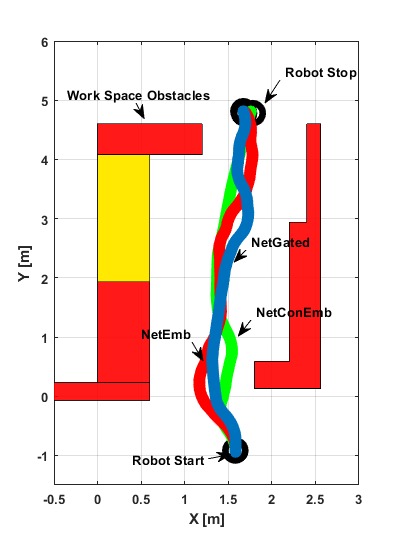}
    \caption{}
  \end{subfigure}
  \hfill
  \begin{subfigure}[b]{0.48\linewidth}
    \includegraphics[keepaspectratio,height=6cm]{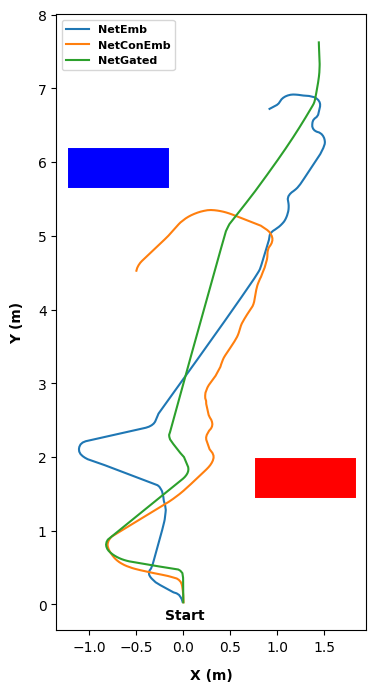}
    \caption{}
  \end{subfigure}
  \caption{Navigation paths using only depth data.}
  \label{fig:paths_depth}
\end{figure}

\begin{figure}
  \centering
  \begin{subfigure}[b]{0.48\linewidth}
    \includegraphics[keepaspectratio,height=6cm]{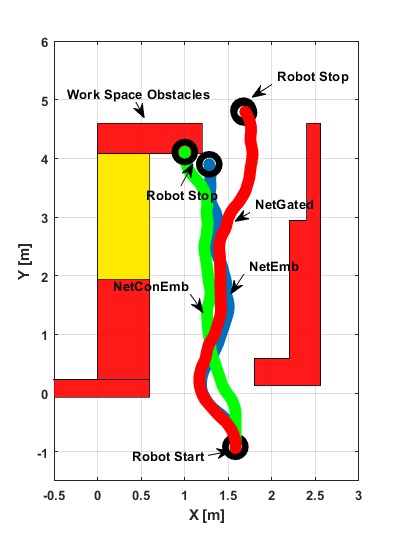}
    \caption{}
  \end{subfigure}
  \hfill
  \begin{subfigure}[b]{0.48\linewidth}
    \includegraphics[keepaspectratio,height=6cm]{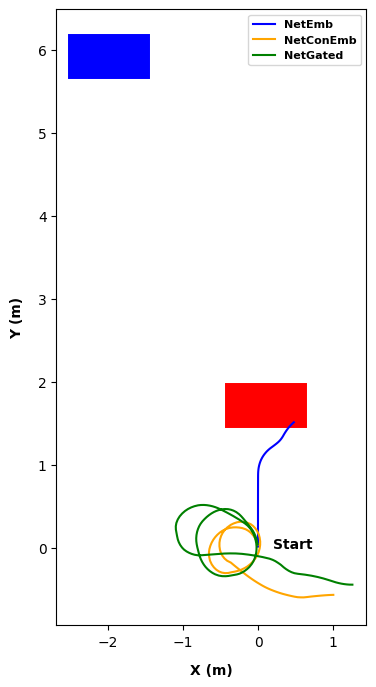}
    \caption{}
  \end{subfigure}
  \caption{Navigation paths using only color data.}
  \label{fig:paths_color}
\end{figure}

To simulate sensor failure in real-world navigation, single-modality information was used to test each model's performance. Reliance on color input alone resulted in suboptimal navigation performance, Fig.~\ref{fig:paths_color}, where all attempts failed except for NetGated, which achieved successful navigation in 3 out of 6 cases (50\%), while depth-only input provided more reliable and stable trajectories as illustrated in Fig.~\ref{fig:paths_depth} with NetGated succeeding in 5 out of 6 cases (83\%) and both NetEmb and NetConEmb achieving a 100\% success rate. This discrepancy highlights the fact that the color data alone does not fully capture the dynamic, noisy, and often unpredictable nature of real-world environments, including light variations. Therefore, while the color modality appears more informative in offline test conditions, depth sensing proves more robust for real-time navigation, reinforcing the importance of sensor fusion in practical robotic systems.

%%%%%%%%%%%%%%%%%%%%%%%%%%%
%%%%%%%%%%%%%%%%%%%%%%%%%%%%%%%%%%

%%%%%%%%%%%%%%%%%%%%%%%%%%%
%%%%%%%%%%%%%%%%%%%%%%%%%%%%%%%%%%
%%%%%%%%%%%%%%%%%%%%%%%%%%%
%%%%%%%%%%%%%%%%%%%%%%%%%%%%%%%%%%

%----------------- Conclusion -------------

\section {Conclusion}

This study introduced a real-time obstacle avoidance system for mobile robots using three end-to-end CNN architectures with RGB-D sensor fusion. The networks were trained to predict low-level steering commands based on synchronized color and depth data captured by an Intel RealSense D415 camera. A comprehensive evaluation using standard regression metrics revealed that NetConEmb achieved the best overall performance, particularly exhibiting the lowest Median Absolute Error. However, NetEmb showed comparable results while offering a substantial reduction in trainable parameters. All models were successfully deployed on a differential-drive mobile robot and validated in both known and unknown environments. The real-world performance of NetConEmb confirmed its robustness and generalization capability, making it a promising solution for vision-based autonomous navigation in various indoor scenarios.

\bibliographystyle{IEEEtran}
\bibliography{references} % Make sure this matches your .bib file name without the extension

\begin{thebibliography}{10}
\providecommand{\url}[1]{#1}
\csname url@samestyle\endcsname
\providecommand{\newblock}{\relax}
\providecommand{\bibinfo}[2]{#2}
\providecommand{\BIBentrySTDinterwordspacing}{\spaceskip=0pt\relax}
\providecommand{\BIBentryALTinterwordstretchfactor}{4}
\providecommand{\BIBentryALTinterwordspacing}{\spaceskip=\fontdimen2\font plus
\BIBentryALTinterwordstretchfactor\fontdimen3\font minus \fontdimen4\font\relax}
\providecommand{\BIBforeignlanguage}[2]{{%
\expandafter\ifx\csname l@#1\endcsname\relax
\typeout{** WARNING: IEEEtran.bst: No hyphenation pattern has been}%
\typeout{** loaded for the language `#1'. Using the pattern for}%
\typeout{** the default language instead.}%
\else
\language=\csname l@#1\endcsname
\fi
#2}}
\providecommand{\BIBdecl}{\relax}
\BIBdecl

\bibitem{xiao2022motion}
X.~Xiao, B.~Liu, G.~Warnell, and P.~Stone, ``Motion planning and control for mobile robot navigation using machine learning: a survey,'' \emph{Autonomous Robots}, pp. 1--29, 2022.

\bibitem{li2021design}
K.~Li and H.~Tu, ``Design and implementation of autonomous mobility algorithm for home service robot based on turtlebot,'' in \emph{2021 IEEE 5th Information Technology, Networking, Electronic and Automation Control Conference (ITNEC)}, vol.~5.\hskip 1em plus 0.5em minus 0.4em\relax IEEE, 2021, pp. 1095--1099.

\bibitem{sepulveda2018deep}
G.~Sepulveda, J.~C. Niebles, and A.~Soto, ``A deep learning based behavioral approach to indoor autonomous navigation,'' in \emph{2018 IEEE International Conference on Robotics and Automation (ICRA)}.\hskip 1em plus 0.5em minus 0.4em\relax IEEE, 2018, pp. 4646--4653.

\bibitem{lu2022online}
Y.~Lu, H.~Wang, N.~Feng, D.~Jiang, and C.~Wei, ``Online interaction method of mobile robot based on single-channel eeg signal and end-to-end cnn with residual block model,'' \emph{Advanced Engineering Informatics}, vol.~52, p. 101595, 2022.

\bibitem{park2020vision}
B.~Park and H.~Oh, ``Vision-based obstacle avoidance for uavs via imitation learning with sequential neural networks,'' \emph{International Journal of Aeronautical and Space Sciences}, vol.~21, no.~3, pp. 768--779, 2020.

\bibitem{lee2021deep}
H.~Y. Lee, H.~W. Ho, and Y.~Zhou, ``Deep learning-based monocular obstacle avoidance for unmanned aerial vehicle navigation in tree plantations: Faster region-based convolutional neural network approach,'' \emph{Journal of Intelligent \& Robotic Systems}, vol. 101, no.~1, p.~5, 2021.

\bibitem{codevilla2018end}
F.~Codevilla, M.~M{\"u}ller, A.~L{\'o}pez, V.~Koltun, and A.~Dosovitskiy, ``End-to-end driving via conditional imitation learning,'' in \emph{2018 IEEE international conference on robotics and automation (ICRA)}.\hskip 1em plus 0.5em minus 0.4em\relax IEEE, 2018, pp. 4693--4700.

\bibitem{9521224}
C.-Y. Tsai, H.~Nisar, and Y.-C. Hu, ``Mapless lidar navigation control of wheeled mobile robots based on deep imitation learning,'' \emph{IEEE Access}, vol.~9, pp. 117\,527--117\,541, 2021.

\bibitem{yan2022mapless}
C.~Yan, J.~Qin, Q.~Liu, Q.~Ma, and Y.~Kang, ``Mapless navigation with safety-enhanced imitation learning,'' \emph{IEEE Transactions on Industrial Electronics}, vol.~70, no.~7, pp. 7073--7081, 2022.

\bibitem{kim2022end}
C.-j. Kim, M.-j. Lee, K.-h. Hwang, and Y.-g. Ha, ``End-to-end deep learning-based autonomous driving control for high-speed environment,'' \emph{The Journal of Supercomputing}, vol.~78, no.~2, pp. 1961--1982, 2022.

\bibitem{nair2020robotic}
R.~S. Nair and P.~Supriya, ``Robotic path planning using recurrent neural networks,'' in \emph{2020 11th International Conference on Computing, Communication and Networking Technologies (ICCCNT)}.\hskip 1em plus 0.5em minus 0.4em\relax IEEE, 2020, pp. 1--5.

\bibitem{bai2018deep}
Z.~Bai, B.~Cai, W.~ShangGuan, and L.~Chai, ``Deep learning based motion planning for autonomous vehicle using spatiotemporal lstm network,'' in \emph{2018 Chinese Automation Congress (CAC)}.\hskip 1em plus 0.5em minus 0.4em\relax IEEE, 2018, pp. 1610--1614.

\bibitem{9144508}
Y.~Zhang, R.~Ge, L.~Lyu, J.~Zhang, C.~Lyu, and X.~Yang, ``A virtual end-to-end learning system for robot navigation based on temporal dependencies,'' \emph{IEEE Access}, vol.~8, pp. 134\,111--134\,123, 2020.

\bibitem{siva2019robot}
S.~Siva, M.~Wigness, J.~Rogers, and H.~Zhang, ``Robot adaptation to unstructured terrains by joint representation and apprenticeship learning,'' in \emph{Robotics: science and systems}, 2019.

\bibitem{zain2025arxiv}
\BIBentryALTinterwordspacing
L.~H. Zain, H.~H. Ammar, and R.~E. Shalaby, ``Imitation learning for obstacle avoidance using end-to-end cnn-based sensor fusion,'' 2025, accepted for presentation at ITC 2025, Cairo, Egypt. [Online]. Available: \url{https://arxiv.org/abs/2507.08112}
\BIBentrySTDinterwordspacing

\bibitem{patel2019deep}
N.~Patel, A.~Choromanska, P.~Krishnamurthy, and F.~Khorrami, ``A deep learning gated architecture for ugv navigation robust to sensor failures,'' \emph{Robotics and Autonomous Systems}, vol. 116, pp. 80--97, 2019.

\end{thebibliography}

\end{document}